\PassOptionsToPackage{unicode}{hyperref}
\PassOptionsToPackage{hyphens}{url}
\PassOptionsToPackage{dvipsnames,svgnames,x11names}{xcolor}
\documentclass[
  12pt,
]{interact}
\usepackage{xcolor}
\usepackage{amsmath,amssymb}
\usepackage{iftex}
\ifPDFTeX
  \usepackage[T1]{fontenc}
  \usepackage[utf8]{inputenc}
  \usepackage{textcomp} 
\else 
  \usepackage{unicode-math} 
  \defaultfontfeatures{Scale=MatchLowercase}
  \defaultfontfeatures[\rmfamily]{Ligatures=TeX,Scale=1}
\fi
\usepackage{lmodern}
\ifPDFTeX\else
\fi
\IfFileExists{upquote.sty}{\usepackage{upquote}}{}
\IfFileExists{microtype.sty}{
  \usepackage[]{microtype}
  \UseMicrotypeSet[protrusion]{basicmath} 
}{}
\usepackage{setspace}
\makeatletter
\@ifundefined{KOMAClassName}{
  \IfFileExists{parskip.sty}{%
    \usepackage{parskip}
  }{
    \setlength{\parindent}{0pt}
    \setlength{\parskip}{6pt plus 2pt minus 1pt}}
}{
  \KOMAoptions{parskip=half}}
\makeatother
\makeatletter
\ifx\paragraph\undefined\else
  \let\oldparagraph\paragraph
  \renewcommand{\paragraph}{
    \@ifstar
      \xxxParagraphStar
      \xxxParagraphNoStar
  }
  \newcommand{\xxxParagraphStar}[1]{\oldparagraph*{#1}\mbox{}}
  \newcommand{\xxxParagraphNoStar}[1]{\oldparagraph{#1}\mbox{}}
\fi
\ifx\subparagraph\undefined\else
  \let\oldsubparagraph\subparagraph
  \renewcommand{\subparagraph}{
    \@ifstar
      \xxxSubParagraphStar
      \xxxSubParagraphNoStar
  }
  \newcommand{\xxxSubParagraphStar}[1]{\oldsubparagraph*{#1}\mbox{}}
  \newcommand{\xxxSubParagraphNoStar}[1]{\oldsubparagraph{#1}\mbox{}}
\fi
\makeatother

\usepackage{color}
\usepackage{fancyvrb}

\DefineVerbatimEnvironment{Highlighting}{Verbatim}{commandchars=\\\{\}}
\usepackage{framed}
\definecolor{shadecolor}{RGB}{241,243,245}
\newenvironment{Shaded}{\begin{snugshade}}{\end{snugshade}}

\newcommand{\AttributeTok}[1]{\textcolor[rgb]{0.40,0.45,0.13}{#1}}

\newcommand{\CharTok}[1]{\textcolor[rgb]{0.13,0.47,0.30}{#1}}
\newcommand{\CommentTok}[1]{\textcolor[rgb]{0.37,0.37,0.37}{#1}}

\newcommand{\ConstantTok}[1]{\textcolor[rgb]{0.56,0.35,0.01}{#1}}

\newcommand{\DecValTok}[1]{\textcolor[rgb]{0.68,0.00,0.00}{#1}}

\newcommand{\FunctionTok}[1]{\textcolor[rgb]{0.28,0.35,0.67}{#1}}

\newcommand{\KeywordTok}[1]{\textcolor[rgb]{0.00,0.23,0.31}{\textbf{#1}}}
\newcommand{\NormalTok}[1]{\textcolor[rgb]{0.00,0.23,0.31}{#1}}

\newcommand{\OtherTok}[1]{\textcolor[rgb]{0.00,0.23,0.31}{#1}}

\newcommand{\SpecialCharTok}[1]{\textcolor[rgb]{0.37,0.37,0.37}{#1}}

\newcommand{\StringTok}[1]{\textcolor[rgb]{0.13,0.47,0.30}{#1}}

\usepackage{longtable,booktabs,array}
\usepackage{calc} 
\usepackage{etoolbox}
\makeatletter
\patchcmd\longtable{\par}{\if@noskipsec\mbox{}\fi\par}{}{}
\makeatother
\IfFileExists{footnotehyper.sty}{\usepackage{footnotehyper}}{\usepackage{footnote}}
\makesavenoteenv{longtable}
\usepackage{graphicx}
\makeatletter
\newsavebox\pandoc@box
\newcommand*\pandocbounded[1]{
  \sbox\pandoc@box{#1}%
  \Gscale@div\@tempa{\textheight}{\dimexpr\ht\pandoc@box+\dp\pandoc@box\relax}%
  \Gscale@div\@tempb{\linewidth}{\wd\pandoc@box}%
  \ifdim\@tempb\p@<\@tempa\p@\let\@tempa\@tempb\fi
  \ifdim\@tempa\p@<\p@\scalebox{\@tempa}{\usebox\pandoc@box}%
  \else\usebox{\pandoc@box}%
  \fi%
}
\def\fps@figure{htbp}
\makeatother

\NewDocumentCommand\citeproctext{}{}

\makeatletter
 \let\@cite@ofmt\@firstofone
 \def\@biblabel#1{}
 \def\@cite#1#2{{#1\if@tempswa , #2\fi}}
\makeatother
\newlength{\cslhangindent}
\newlength{\csllabelwidth}
\newenvironment{CSLReferences}[2] 
 {\begin{list}{}{%
  \setlength{\itemindent}{0pt}
  \setlength{\leftmargin}{0pt}
  \setlength{\parsep}{0pt}
  \ifodd #1
   \setlength{\leftmargin}{\cslhangindent}
   \setlength{\itemindent}{-1\cslhangindent}
  \fi
  \setlength{\itemsep}{#2\baselineskip}}}
 {\end{list}}
\usepackage{calc}

\providecommand{\tightlist}{%
  \setlength{\itemsep}{0pt}\setlength{\parskip}{0pt}}

\usepackage{amsmath}
\usepackage{amssymb}
\usepackage{fvextra}
\DefineVerbatimEnvironment{verbatim}{Verbatim}{breaklines=true, breakanywhere=true, breakindent=1em, fontsize=\small}
\DefineVerbatimEnvironment{Highlighting}{Verbatim}{commandchars=\\\{\}, breaklines=true, breakanywhere=true, breakindent=1em, fontsize=\small}
\makeatletter
\AtBeginDocument{%
  \captionsetup[codelisting]{skip=-0.8em}%
  \renewenvironment{codelisting}{%
    \def\@captype{codelisting}%
    \par\addvspace{0.5em}\noindent
  }{%
    \par\addvspace{0.5em}%
  }%
}
\makeatother
\AtBeginDocument{\floatplacement{codelisting}{H}}
\usepackage{orcidlink}
\makeatletter
\@ifpackageloaded{caption}{}{\usepackage{caption}}
\AtBeginDocument{%
\ifdefined\contentsname
  \renewcommand*\contentsname{Table of contents}
\else
  \newcommand\contentsname{Table of contents}
\fi
\ifdefined\listfigurename
  \renewcommand*\listfigurename{List of Figures}
\else
  \newcommand\listfigurename{List of Figures}
\fi
\ifdefined\listtablename
  \renewcommand*\listtablename{List of Tables}
\else
  \newcommand\listtablename{List of Tables}
\fi
\ifdefined\figurename
  \renewcommand*\figurename{Figure}
\else
  \newcommand\figurename{Figure}
\fi
\ifdefined\tablename
  \renewcommand*\tablename{Table}
\else
  \newcommand\tablename{Table}
\fi
}
\@ifpackageloaded{float}{}{\usepackage{float}}
\floatstyle{ruled}
\@ifundefined{c@chapter}{\newfloat{codelisting}{h}{lop}}{\newfloat{codelisting}{h}{lop}[chapter]}
\floatname{codelisting}{Listing}

\makeatother
\makeatletter
\@ifpackageloaded{caption}{}{\usepackage{caption}}
\@ifpackageloaded{subcaption}{}{\usepackage{subcaption}}
\makeatother
\usepackage{bookmark}
\IfFileExists{xurl.sty}{\usepackage{xurl}}{} 
\hypersetup{
  pdftitle={Code availability is not enough: schemas for documenting and extending data analysis},
  pdfauthor={H. Sherry Zhang; Roger D. Peng},
  pdfkeywords={Reproducibility, Large Language Models (LLMs), Data
analysis decisions, Many-analysts study, Statistical workflow},
  colorlinks=true,
  linkcolor={blue},
  filecolor={Maroon},
  citecolor={Blue},
  urlcolor={Blue},
  pdfcreator={LaTeX via pandoc}}

\title{Code availability is not enough: schemas for documenting and
extending data analysis}
\author{H. Sherry Zhang$\textsuperscript{1}$, Roger D.
Peng$\textsuperscript{1}$}

\thanks{CONTACT: H. Sherry
Zhang. Email: \href{mailto:hsherryzhang@utexas.edu}{\nolinkurl{hsherryzhang@utexas.edu}}. }
\begin{document}
\captionsetup{labelsep=space}
\maketitle
\textsuperscript{1} Department of Statistics and Data
Science, University of Texas at Austin, Texas, United States
\begin{abstract}
A data analysis involves a sequence of decisions, from data cleaning to
model specification, each of which can affect the reported result.
Reproducing that result is not the same as understanding the sequence of
decisions that produced it and the reasons for each: the code mixes
those decisions with implementation details, while the manuscript states
the major decisions and their rationale but scatters them across its
sections and often leaves out the minor ones. A reader may then ask how
the result would change under a different decision: the question itself
is easy to pose, but implementing it is often non-trivial in practice.
In this work, we propose a schema that represents a data analysis as a
collection of decisions in natural language, from which large language
models (LLMs) can generate and validate analysis scripts directly. This
workflow can also expand a single analysis into a full multiverse of
alternatives, specified in an alternative file, ready for comparison. We
demonstrate the proposed workflow in two applications. In the first, we
construct a schema for a published regression analysis of heaped smoking
data and show that it reproduces the reported estimates, and that a
reviewer-suggested alternative can be specified through the alternative
file and turned into code automatically. In the second, where no schema
is provided, we show how the workflow serves instead as a
reproducibility check, applied to the many-analysts football study. As
an intermediate representation between the manuscript and code scripts,
our schema gives a self-contained account of an analysis and makes it
extensible from a reader's perspective.
\end{abstract}
\begin{keywords}
\def\sep{;\ }
Reproducibility\sep Large Language Models (LLMs)\sep Data analysis
decisions\sep Many-analysts study\sep 
Statistical workflow
\end{keywords}

\setstretch{1.15}
\newpage

\section{Introduction}\label{sec-intro}

A data analysis is the product of a chain of decisions, from how the
data are cleaned to how the model is specified. Those decisions are what
a reader needs in order to comprehend an analysis, judge whether it is
reasonable, and extend it by asking how the result would change under
different choices. Standard practice in reproducible research (Peng,
2011; Wrobel et al., 2024) asks for the code and the data to be made
available, but does not extend to the decisions behind them.

This gap arises because code and the decisions it implements operate at
different levels: code specifies precisely what a computer should do,
while a decision has a contextual interpretation that the code itself
does not capture. Consider a line such as
\texttt{dplyr::filter(df,\ x\ \textgreater{}\ 0)}, which has a precise
meaning as code: keep rows where \texttt{x} is positive. Which decision
this line encodes is not something the code can answer: it could
represent restricting the analysis to a subpopulation for which
\(x > 0\) is substantively meaningful, or excluding observations flagged
as invalid by an encoding convention where \(x \le 0\) denotes
missingness, and nothing in the \texttt{filter()} syntax indicates which
is intended. Nor does code generally carry the rationale behind a
decision, unless the author has left detailed comments. A better place
to state the decisions is the manuscript, where they are described along
with their rationale. But this information is scattered across the
manuscript rather than collected in one place, with preprocessing and
modeling decisions appearing in different sections. Authors may also
simply leave out decisions they consider minor, on the assumption that
the code provides all the detail a reader would need. Reconstructing the
full sequence, therefore, means searching across the paper and the code
for decisions and their rationale before a reader can begin to
comprehend the analysis.

Literate programming tools such as Sweave (Leisch, 2002), R Markdown
(Xie et al., 2018), Jupyter notebooks (Kluyver et al., 2016), and Quarto
(Allaire et al., 2026) integrate narrative and code within a single
document, yet a published notebook does not resolve this problem.
Substantive decisions, such as the exclusion of observations or the
choice of covariates, remain interleaved with routine data manipulation,
such as the renaming of columns, and a reader needs to examine the code
to distinguish one from the other. Moreover, the narrative exists to
explain the code, and the format places no requirement on what it must
cover. An author narrates the steps they consider worth explaining, not
the complete set of decisions needed to obtain the result. Ideally, a
reader would understand an analysis by following its decisions step by
step to reach the result, and would consult the code only when the
implementation of a particular decision is of interest. What this
requires is a self-contained record of the decisions and their
rationale. With such a record, a reader can assess each decision that
leads to the result and raise alternatives to investigate how the result
changes. Once the chain of decisions has been specified, its
implementation in code reduces largely to a translation task, which
large language models (LLMs) now perform with reasonable reliability
(Chen et al., 2021; Huang et al., 2022; Jiang et al., 2026; Lai et al.,
2023; Shah et al., 2026).

In this paper, we present a schema that records a data analysis as a
sequence of decisions, each stated with the objective it addresses and
the rationale for choosing it, together with the variables it consumes
and produces. We also demonstrate how this schema, coupled with LLMs,
can be used to generate code scripts from natural-language descriptions
in manuscripts and to explore alternative decisions. We present two
applications. The first takes the perspective of an author supplying a
schema alongside a published manuscript, using the GAITD regression
analysis of heaped smoking data in Yee et al. (2025). A reader can then
specify an alternative file, coupled with the original schema, to
compare model performance under a different inflation-alteration
specification. The second, the many-analysts football study, considers
the opposite situation, where no schema is supplied and the workflow
serves instead as a reproducibility test. It extracts a schema from a
simplified version of each team's report, generates and validates the
corresponding code, and compares the result against that team's
published estimate.

The paper is organized as follows: In Section~\ref{sec-method}, we
present our methodology, including the schema representation for
describing a data analysis, the specification of the alternative file,
and the workflow for generating and validating code from the schema with
LLMs. In Section~\ref{sec-gaitd}, we construct a schema for a published
GAITD regression analysis and show how it allows a reader to evaluate an
alternative specification directly. In Section~\ref{sec-application}, we
apply the workflow as a reproducibility test to the many-analysts
football study, recovering the reported results for most teams and
discussing the failed cases. Finally, in Section~\ref{sec-discussion},
we discuss the implications of our work and potential future directions.

\section{Methodology}\label{sec-method}

\subsection{A schema for describing data analysis}\label{sec-schema}

We represent a data analysis,
\(\mathcal{S} = \{A_1, A_2, \cdots, A_n\}\), as a sequence of \(n\)
analysis decisions \(A_i\), applied to a dataset \(\mathcal{D}\), where
each step, \(A_i\), is defined as a tuple:

\begin{equation}
A_{i} = \langle f_{i}, \pi_{i}, \rho_{i}, \mathcal{I}_{i}, \mathcal{O}_{i}, \omega_{i} \rangle.
\end{equation}

\(f_{i} \in \mathcal{F}\) is the \emph{objective} of a step, identifying
the problem the \emph{decision}, \(\pi_{i} \in \Pi\), is trying to
solve. \(\rho_{i}\) is the \emph{rationale} provided for choosing the
decision \(\pi_{i}\). All three fields are written in short
natural-language sentences, and together they form a self-contained
account of each decision in the analysis. As an example,
Listing~\ref{lst-schema} presents the schema that describes the steps
for calculating the Human Development Index (HDI) (United Nations
Development Programme, 2025):

\begin{enumerate}
\def\labelenumi{\arabic{enumi})}
\tightlist
\item
  Scale all four variables (life expectancy, expected years of
  schooling, mean years of schooling, and GNI per capita) to the {[}0,
  1{]} range.
\item
  Combine the two education variables (expected years of schooling and
  mean years of schooling) into a single education variable.
\item
  Combine the three dimensions (health, education, and income) into a
  single index using a geometric mean.
\end{enumerate}

\begin{codelisting}

\caption{\label{lst-schema}An example schema for calculating the Human
Development Index (HDI).}

\centering{

\begin{Shaded}
\begin{Highlighting}[]
\FunctionTok{meta}\KeywordTok{:}
\AttributeTok{  }\FunctionTok{type}\KeywordTok{:}\AttributeTok{ schema}
\FunctionTok{nodes}\KeywordTok{:}
\KeywordTok{{-}}\AttributeTok{ }\FunctionTok{id}\KeywordTok{:}\AttributeTok{ step{-}scaling}
\AttributeTok{  }\FunctionTok{objective}\KeywordTok{:}\AttributeTok{ how to handle variables in different scales?}
\AttributeTok{  }\FunctionTok{decision}\KeywordTok{:}\AttributeTok{ apply min{-}max scaling to each variable}
\AttributeTok{  }\FunctionTok{rationale}\KeywordTok{:}\AttributeTok{ to convert the variables to the same scale for future}
\AttributeTok{    construction of a single index}
\KeywordTok{{-}}\AttributeTok{ }\FunctionTok{id}\KeywordTok{:}\AttributeTok{ step{-}education}
\AttributeTok{  }\FunctionTok{objective}\KeywordTok{:}\AttributeTok{ how to combine the two education{-}related variables?}
\AttributeTok{  }\FunctionTok{decision}\KeywordTok{:}\AttributeTok{ take the average}
\AttributeTok{  }\FunctionTok{rationale}\KeywordTok{:}\AttributeTok{ the arithmetic mean allows perfect substitutability}
\AttributeTok{    between the two variables}
\KeywordTok{{-}}\AttributeTok{ }\FunctionTok{id}\KeywordTok{:}\AttributeTok{ step{-}combine}
\AttributeTok{  }\FunctionTok{objective}\KeywordTok{:}\AttributeTok{ how to combine the three dimensions into a single index?}
\AttributeTok{  }\FunctionTok{decision}\KeywordTok{:}\AttributeTok{ use the geometric mean}
\AttributeTok{  }\FunctionTok{rationale}\KeywordTok{:}\AttributeTok{ the geometric mean penalizes uneven development }
\AttributeTok{    among the three core dimensions}
\end{Highlighting}
\end{Shaded}

}

\end{codelisting}%

In the file, the \texttt{meta} field at the top marks the document as a
schema (\texttt{type:\ schema}), and \texttt{nodes} below it records the
analysis as a list of steps. Each step carries an \texttt{id} that names
it, together with the \texttt{objective}, \texttt{decision}, and
\texttt{rationale} fields holding \(f_i\), \(\pi_i\), and \(\rho_i\).
For instance, the objective of step 3 is to combine the three dimensions
to create a single index. The geometric mean (the decision) is chosen
over a simple arithmetic mean (an alternative, introduced in
Section~\ref{sec-alternative}) because it penalizes uneven development
among the three dimensions (the rationale).

Generating code from the schema also requires each step to be tied to
concrete variables in the data. Where a dataset \(\mathcal{D}\) is
available, \(\mathcal{I}_i\) and \(\mathcal{O}_i\) record that mapping
as part of the schema, so that it is fixed in advance rather than
inferred at the point of code generation. Because the schema specifies
an analysis before its code exists, \(\mathcal{O}_i\) may name a
variable that is not yet in the dataset. In contrast, \(\mathcal{I}_i\)
must already be in scope, comprising variables in the original dataset
together with the outputs of all previous steps:

\[\mathcal{I}_{i} \subseteq \left( \mathcal{D} \cup \bigcup_{j < i} \mathcal{O}_{j} \right), \quad \forall i \in \{2, \dots, n\}.\]

A schema can be written by hand, or drafted by an LLM from an existing
description of the analysis, such as the methodology section of a
manuscript. In the latter case, \(\mathcal{I}_i\) and \(\mathcal{O}_i\)
are inferred by the model whenever a dataset is provided, and each step
carries a categorical confidence score, \(\omega_{i}\) (low, medium,
high), defined as the model's confidence that the recorded
\((f_i, \pi_i, \rho_i)\) correctly reflects the source text.

\subsection{Describing alternatives}\label{sec-alternative}

Recording a step's decision \(\pi_i\) does not mean it is the only one
that could satisfy its objective \(f_i\): a reader may want to revisit
the decision made for some of these steps and consider alternatives that
address the same objective, the premise of multiverse analysis
(Simonsohn et al., 2020; Steegen et al., 2016). Formally, a step in the
schema can have a candidate set of alternatives: step \(i\) may admit
\(k_i\) decisions that all satisfy its objective \(f_i\),
\(\{A_i^1, \dots, A_i^{k_i}\}\), of which the decision \(\pi_i\)
recorded in the schema is just one member, \(A_i^1\). An alternative
file specifies the remaining candidates, \(A_i^2, \dots, A_i^{k_i}\),
for a subset of steps \(C \subseteq \{1, \dots, n\}\).

The overall structure of an alternative file is shown in
Listing~\ref{lst-alternatives-skeleton}. The \texttt{meta} field at the
top identifies the file as an alternative specification
(\texttt{type:\ alternatives}), and the file-level \texttt{branch}
field, either \texttt{single} or \texttt{multi}, determines how the
candidates at each step are combined. The \texttt{nodes} field then
holds one entry for every step in \(C\). Each entry uses
\texttt{overrides} to name the step it applies to in the original
schema, then lists that step's alternative decisions and their
rationale.

\begin{codelisting}

\caption{\label{lst-alternatives-skeleton}The skeleton of an alternative
YAML file, which consists of a \texttt{branch} mode and a list of
covered steps.}

\centering{

\begin{Shaded}
\begin{Highlighting}[]
\FunctionTok{meta}\KeywordTok{:}
\AttributeTok{  }\FunctionTok{type}\KeywordTok{:}\AttributeTok{ alternatives}
\AttributeTok{  }\FunctionTok{branch}\KeywordTok{:}\AttributeTok{ single | multi}
\FunctionTok{nodes}\KeywordTok{:}
\AttributeTok{  }\KeywordTok{{-}}\AttributeTok{ }\FunctionTok{overrides}\KeywordTok{:}\AttributeTok{ \textless{}step id\textgreater{}}
\AttributeTok{    }\FunctionTok{alternatives}\KeywordTok{:}
\AttributeTok{      }\KeywordTok{{-}}\AttributeTok{ }\FunctionTok{id}\KeywordTok{:}\AttributeTok{ \textless{}alternative id\textgreater{}}
\FunctionTok{        decision}\KeywordTok{: }\AttributeTok{\textless{}...}\CharTok{\textgreater{}}
\FunctionTok{        rationale}\KeywordTok{: }\AttributeTok{\textless{}...}\CharTok{\textgreater{}}
\AttributeTok{      }\KeywordTok{{-}}\AttributeTok{ }\FunctionTok{id}\KeywordTok{:}\AttributeTok{ \textless{}alternative id\textgreater{}}
\AttributeTok{        ...}
\AttributeTok{  }\KeywordTok{{-}}\AttributeTok{ }\FunctionTok{overrides}\KeywordTok{:}\AttributeTok{ \textless{}step id\textgreater{}}
\AttributeTok{    }\FunctionTok{alternatives}\KeywordTok{:}
\AttributeTok{      ...}
\AttributeTok{  ...}
\end{Highlighting}
\end{Shaded}

}

\end{codelisting}%

Under \texttt{branch:\ single}, every step is restricted to a single
candidate decision. These alternatives, together with the rest of the
original schema, form one new analysis. This is useful when an analyst
already has a specific alternative analysis in mind and wants to
evaluate it as a whole. Listing~\ref{lst-alternatives-single} fills in
the skeleton to construct an alternative HDI computation, using a
harmonic mean to combine the two education variables in step 2 and an
arithmetic mean to combine the three dimensions in step 3.

\begin{codelisting}

\caption{\label{lst-alternatives-single}A sample alternative file that
changes the decisions for steps 2 and 3 in calculating the Human
Development Index (HDI).}

\centering{

\begin{Shaded}
\begin{Highlighting}[]
\FunctionTok{meta}\KeywordTok{:}
\AttributeTok{  }\FunctionTok{type}\KeywordTok{:}\AttributeTok{ alternatives}
\AttributeTok{  }\FunctionTok{branch}\KeywordTok{:}\AttributeTok{ single}
\FunctionTok{nodes}\KeywordTok{:}
\AttributeTok{  }\KeywordTok{{-}}\AttributeTok{ }\FunctionTok{overrides}\KeywordTok{:}\AttributeTok{ step{-}education}
\AttributeTok{    }\FunctionTok{alternatives}\KeywordTok{:}
\AttributeTok{      }\KeywordTok{{-}}\AttributeTok{ }\FunctionTok{id}\KeywordTok{:}\AttributeTok{ harmonic{-}mean}
\AttributeTok{        }\FunctionTok{decision}\KeywordTok{:}\AttributeTok{ take the harmonic mean of the two education variables}
\AttributeTok{        }\FunctionTok{rationale}\KeywordTok{:}\AttributeTok{ penalizes disparity between the two schooling measures}
\AttributeTok{          rather than allowing one to compensate for the other}
\AttributeTok{  }\KeywordTok{{-}}\AttributeTok{ }\FunctionTok{overrides}\KeywordTok{:}\AttributeTok{ step{-}combine}
\AttributeTok{    }\FunctionTok{alternatives}\KeywordTok{:}
\AttributeTok{      }\KeywordTok{{-}}\AttributeTok{ }\FunctionTok{id}\KeywordTok{:}\AttributeTok{ arith{-}mean}
\AttributeTok{        }\FunctionTok{decision}\KeywordTok{:}\AttributeTok{ use an arithmetic mean}
\AttributeTok{        }\FunctionTok{rationale}\KeywordTok{:}\AttributeTok{ a simple average of the three dimensions}
\end{Highlighting}
\end{Shaded}

}

\end{codelisting}%

Sometimes a reader instead wants to test a range of alternatives for one
or more steps independently. Under \texttt{branch:\ multi}, each covered
step may take any of its alternatives or keep its original decision, and
every such combination across steps becomes a separate analysis. In
Listing~\ref{lst-alternatives-multi}, one alternative is specified for
the education step and two for the combination step. Together with the
original schema, this produces \((1+1) \times (2+1) = 6\) analyses. More
generally, with \(k_i\) candidates at each step \(i \in C\), crossing
all candidates yields \(\prod_{i \in C} k_i\) analyses.

\begin{codelisting}

\caption{\label{lst-alternatives-multi}A sample alternative file that
explores one alternative decision for step 2 and two for step 3 in
calculating the Human Development Index (HDI). Combining these
alternatives with the original schema produces a total of (1+1) × (2+1)
= 6 analyses.}

\centering{

\begin{Shaded}
\begin{Highlighting}[]
\FunctionTok{meta}\KeywordTok{:}
\AttributeTok{  }\FunctionTok{type}\KeywordTok{:}\AttributeTok{ alternatives}
\AttributeTok{  }\FunctionTok{branch}\KeywordTok{:}\AttributeTok{ multi}
\FunctionTok{nodes}\KeywordTok{:}
\AttributeTok{  }\KeywordTok{{-}}\AttributeTok{ }\FunctionTok{overrides}\KeywordTok{:}\AttributeTok{ step{-}education}
\AttributeTok{    }\FunctionTok{alternatives}\KeywordTok{:}
\AttributeTok{      }\KeywordTok{{-}}\AttributeTok{ }\FunctionTok{id}\KeywordTok{:}\AttributeTok{ harmonic{-}mean}
\AttributeTok{        }\FunctionTok{decision}\KeywordTok{:}\AttributeTok{ take the harmonic mean of the two education variables}
\AttributeTok{        }\FunctionTok{rationale}\KeywordTok{:}\AttributeTok{ penalizes disparity between the two schooling measures}
\AttributeTok{          rather than allowing one to compensate for the other}
\AttributeTok{  }\KeywordTok{{-}}\AttributeTok{ }\FunctionTok{overrides}\KeywordTok{:}\AttributeTok{ step{-}combine}
\AttributeTok{    }\FunctionTok{alternatives}\KeywordTok{:}
\AttributeTok{      }\KeywordTok{{-}}\AttributeTok{ }\FunctionTok{id}\KeywordTok{:}\AttributeTok{ arith{-}mean}
\AttributeTok{        }\FunctionTok{decision}\KeywordTok{:}\AttributeTok{ use an arithmetic mean}
\AttributeTok{        }\FunctionTok{rationale}\KeywordTok{:}\AttributeTok{ a simple average of the three dimensions}
\AttributeTok{      }\KeywordTok{{-}}\AttributeTok{ }\FunctionTok{id}\KeywordTok{:}\AttributeTok{ min{-}agg}
\AttributeTok{        }\FunctionTok{decision}\KeywordTok{:}\AttributeTok{ use the minimum of the three dimensions}
\AttributeTok{        }\FunctionTok{rationale}\KeywordTok{:}\AttributeTok{ a "weakest{-}link" aggregation that penalizes any single}
\AttributeTok{          lagging dimension more strongly than an average would}
\end{Highlighting}
\end{Shaded}

}

\end{codelisting}%

Existing multiverse tooling, such as the \texttt{multiverse} package in
R (Sarma et al., 2023), expresses each branch in code, while our
alternatives are instead specified at the level of decisions, in natural
language, from which modern LLMs can generate the corresponding code.
Keeping them in a separate file also changes who can pose the
alternative question: a multiverse is conventionally built by the
original analyst exploring their own garden of forking paths, whereas an
alternative file can be written by a reader after reading the published
manuscript, against a schema the author has already supplied, to
investigate questions that interest them.

\subsection{Workflow}\label{sec-workflow}

The reliability of the generated code depends primarily on the quality
of the schema. A schema that faithfully records the decisions of an
analysis, in the form set out in Section~\ref{sec-schema}, leaves little
to be interpreted at the code generation stage, since the objective and
the decision at each step are stated explicitly. The errors that remain
are in the code implementation rather than the analysis: a missing
\texttt{library()} call, or a misnamed function argument.

The validation stage is designed to resolve errors of this kind. The
validation agent executes the script in a sandbox environment and, when
execution fails, receives the script together with the error message and
the line number at which it occurred, and returns a revised script. The
revised script is then executed again, and the cycle continues until the
script completes without error, a maximum number of attempts is reached,
or the user terminates the process. Application to the many-analysts
football study in Section~\ref{sec-application} shows that this
procedure resolves most of the syntax errors that arise in data
wrangling and statistical modeling.

The same generation-validation procedure applies to multiple analyses.
When a schema is augmented with an alternative file, the expansion
described in Section~\ref{sec-alternative} produces one schema for each
combination of decisions, and each resulting schema is passed through
the same generation and validation procedure independently.

\begin{figure}

\centering{

\includegraphics[width=1\linewidth,height=\textheight,keepaspectratio]{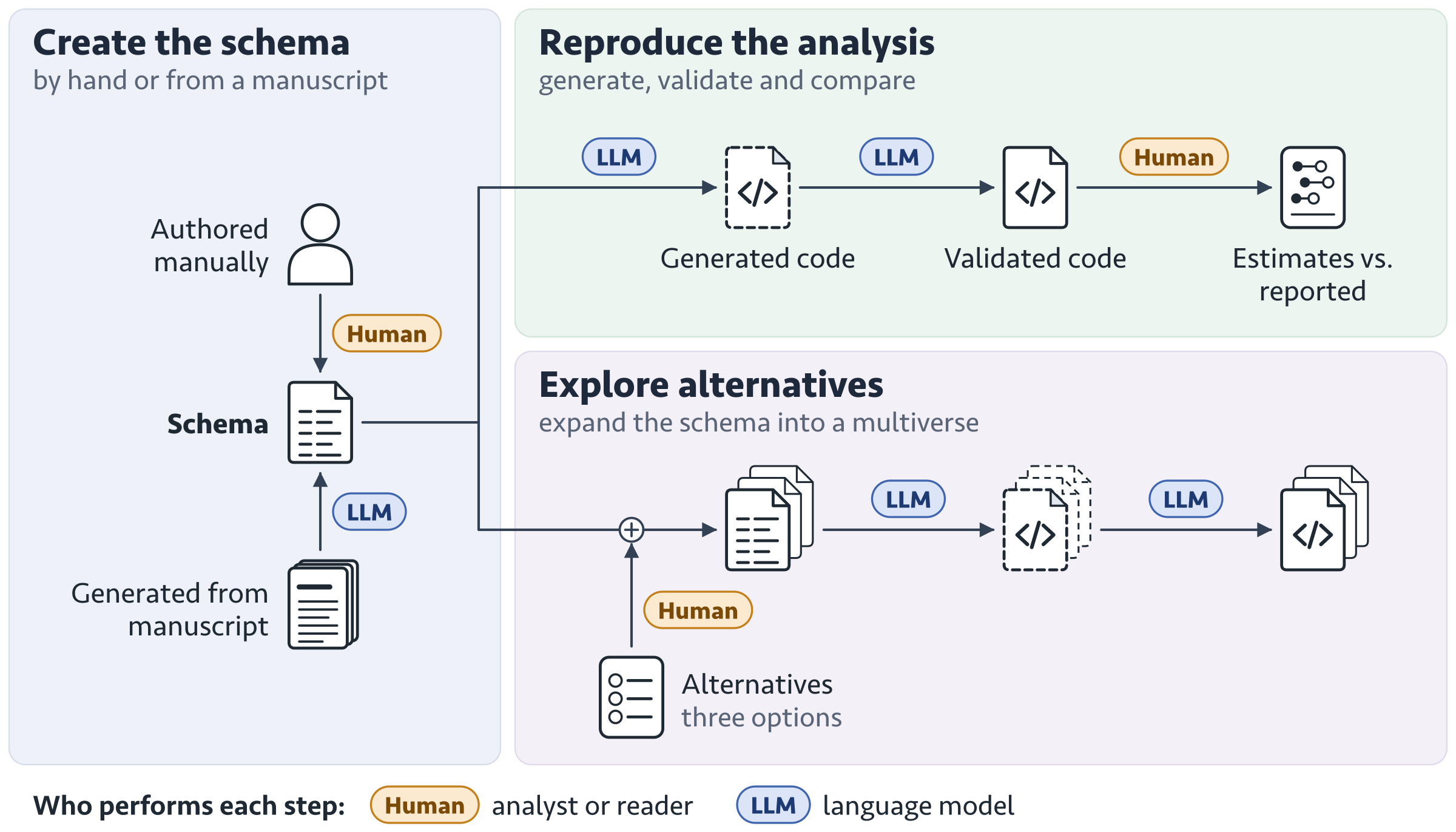}

}

\caption{\label{fig-workflow}Overview of the schema-driven workflow. A
schema is either authored manually or extracted from a manuscript by an
LLM (left). To reproduce the analysis (top), an LLM generates code from
the schema and validates it through iterative execution and repair in a
sandboxed environment, and the analyst runs the validated script and
compares its estimates with those reported. To explore alternatives
(bottom), an alternative file is merged with the schema to give one
schema per alternative, each passed through the same generation and
validation. Badges mark whether a step is performed by a human or an
LLM.}

\end{figure}

Figure~\ref{fig-workflow} gives an overview of the workflow, which
provides two paths for using a schema. Passing a schema through the code
generation and validation stages produces a script whose execution
yields the estimates of the analysis, which can be compared with the
estimates originally reported, serving as a reproducibility check. A
schema can also be augmented with an alternative file, expanding it into
a multiverse of parallel analyses, so that the author and subsequent
readers can explore the reported result under different decisions.

\section{Application: GAITD regression of heaped smoking
data}\label{sec-gaitd}

In this section, we demonstrate the workflow in the case where an author
provides a faithful schema alongside a published manuscript, which a
reader can then use to question and explore the analysis, following the
lower path in Figure~\ref{fig-workflow}. We use Yee et al. (2025)'s
GAITD regression analysis of heaped smoking data, published in the
\emph{Annals of Applied Statistics}, as the example. Yee et al. (2025)
report four separate smoking-related responses, and for demonstration
purposes, we build the schema for only the first response, smoking
initiation age (P1, Section 5.1 of their paper). The same workflow
extends to the other three cases.

The data come from the smoking questionnaire of the 2017--2020 cycle of
the National Health and Nutrition Examination Survey (NHANES), available
as \texttt{smqP} in the \texttt{VGAMdata} package (Yee \& Gray, 2025).
Each row is a survey respondent, recording their smoking history
(initiation age \texttt{SIA}, cessation age \texttt{SCA}, and daily
cigarette consumption, \texttt{TC30} for current smokers and
\texttt{TCquit} for former smokers), their age, and the standard
demographic covariates in smoking studies: gender, race, marital status,
and education. Several of these variables are missing by design:
\texttt{SCA} and \texttt{TCquit} exist only for former smokers, and
education and marital status were not collected for respondents under
20; age is also top-coded at 80. The response, \texttt{SIA}, is a count
whose distribution is heaped: certain ages are reported far more often
than a classic count distribution would predict. GAITD (generally
altered, inflated, truncated, and deflated) regression accommodates such
heaping by modifying a parent count distribution, such as the Poisson or
negative binomial, at specified values, which may receive additional
probability (inflation), reduced probability (deflation), probabilities
modeled separately from the parent (alteration), or no probability at
all (truncation). Which ages to treat as special, and which operation to
apply (inflation, deflation, alteration, or truncation) to each, are
modeling decisions typically guided by an exploratory spikeplot of the
empirical distribution of \texttt{SIA}.

The schema in Listing~\ref{lst-heaping} comprises fourteen entries, each
recording an \emph{objective}, a \emph{decision}, and a \emph{rationale}
in the structure introduced in Section~\ref{sec-schema}. Consider
\texttt{special\_value\_mid\_risk}: the decision is to treat ages 12--14
and 19--21 as one inflated group with extra probability mass, since
these ages show a moderate, secondary peak in the spikeplots, with the
high-risk group (ages 15--18) showing the most prominent peak. Recording
the rationale along with the decision allows a reader to agree or
disagree with the authors' choice, and to question and explore the
analysis, as we do later in this section.

\begin{codelisting}

\caption{\label{lst-heaping}A schema for the GAITD regression analysis
of smoking initiation age in Yee et al. (2025), constructed from the
manuscript and intended to be provided alongside it as a self-contained
account of the analysis.}

\centering{

\begin{Shaded}
\begin{Highlighting}[]
\FunctionTok{meta}\KeywordTok{:}
\AttributeTok{  }\FunctionTok{type}\KeywordTok{:}\AttributeTok{ schema}
\FunctionTok{nodes}\KeywordTok{:}
\KeywordTok{{-}}\AttributeTok{ }\FunctionTok{id}\KeywordTok{:}\AttributeTok{ complete\_case\_filtering}
\AttributeTok{  }\FunctionTok{objective}\KeywordTok{:}\AttributeTok{ handle missing data in selected variables}
\AttributeTok{  }\FunctionTok{decision}\KeywordTok{:}\AttributeTok{ apply complete{-}case filtering to all columns except TC30, TCquit, and SCA which have NAs by design}
\AttributeTok{  }\FunctionTok{rationale}\KeywordTok{:}\AttributeTok{ TC30, TCquit, and SCA are exempted from the completeness check since they are interlaced with NAs by design (e.g., SCA being NA means the person has not yet quit, a valid state rather than missingness); every other selected column is required to be non{-}missing because most smoking studies adjust for age, marital status, education and race, and this analysis follows suit. }

\KeywordTok{{-}}\AttributeTok{ }\FunctionTok{id}\KeywordTok{:}\AttributeTok{ age\_restriction}
\AttributeTok{  }\FunctionTok{objective}\KeywordTok{:}\AttributeTok{ define the age range for analysis}
\AttributeTok{  }\FunctionTok{decision}\KeywordTok{:}\AttributeTok{ restrict analysis to respondents aged less than 80 years (age \textless{} 80)}
\AttributeTok{  }\FunctionTok{rationale}\KeywordTok{:}\AttributeTok{ age was right{-}censored at 80}

\KeywordTok{{-}}\AttributeTok{ }\FunctionTok{id}\KeywordTok{:}\AttributeTok{ data\_quality\_correction}
\AttributeTok{  }\FunctionTok{objective}\KeywordTok{:}\AttributeTok{ handle logically impossible self{-}reported values}
\AttributeTok{  }\FunctionTok{decision}\KeywordTok{:}\AttributeTok{ set both SCA and SIA to NA where SCA {-} SIA \textless{} 0}
\AttributeTok{  }\FunctionTok{rationale}\KeywordTok{:}\AttributeTok{ quitting age before smoking initiation age is a logical impossibility and such values cannot be trusted}

\KeywordTok{{-}}\AttributeTok{ }\FunctionTok{id}\KeywordTok{:}\AttributeTok{ impute\_marital\_young}
\AttributeTok{  }\FunctionTok{objective}\KeywordTok{:}\AttributeTok{ assign a defensible marital status for young respondents, regardless of what was recorded}
\AttributeTok{  }\FunctionTok{decision}\KeywordTok{:}\AttributeTok{ recode marital status to "Never married" for everyone age \textless{} 20}
\AttributeTok{  }\FunctionTok{rationale}\KeywordTok{:}\AttributeTok{ this age group is effectively certain to have never been married, so the value is set by age rather than trusting the raw survey response}

\KeywordTok{{-}}\AttributeTok{ }\FunctionTok{id}\KeywordTok{:}\AttributeTok{ impute\_education\_young}
\AttributeTok{  }\FunctionTok{objective}\KeywordTok{:}\AttributeTok{ assign a defensible education category for young respondents, regardless of what was recorded}
\AttributeTok{  }\FunctionTok{decision}\KeywordTok{:}\AttributeTok{ recode education to "\textless{} 9th grade" for age \textless{} 14; "9{-}11th grade" for 14 \textless{}= age \textless{}= 16; "High school" for 17 \textless{}= age \textless{}= 19}
\AttributeTok{  }\FunctionTok{rationale}\KeywordTok{:}\AttributeTok{ for respondents still of school age, highest{-}grade{-}completed is essentially determined by age, so the value is set by age rather than trusting the raw survey response}

\KeywordTok{{-}}\AttributeTok{ }\FunctionTok{id}\KeywordTok{:}\AttributeTok{ define\_ever\_smokers\_sample}
\AttributeTok{  }\FunctionTok{objective}\KeywordTok{:}\AttributeTok{ define the analytic sample of ever{-}smokers}
\AttributeTok{  }\FunctionTok{decision}\KeywordTok{:}\AttributeTok{ include as valid observation if both conditions are met {-}{-} 1) SIA is not missing, and 2) either SCA and TCquit are both not missing, or SCA is missing and TC30 is not missing}
\AttributeTok{  }\FunctionTok{rationale}\KeywordTok{:}\AttributeTok{ SCA is missing by design for anyone who has not yet quit. the two{-}branch rule instead admits ex{-}smokers through the SCA/TCquit branch and current smokers through the TC30 branch}

\KeywordTok{{-}}\AttributeTok{ }\FunctionTok{id}\KeywordTok{:}\AttributeTok{ parent\_distribution\_choice}
\AttributeTok{  }\FunctionTok{objective}\KeywordTok{:}\AttributeTok{ select appropriate probability distribution for count response variable}
\AttributeTok{  }\FunctionTok{decision}\KeywordTok{:}\AttributeTok{ adopt negative binomial distribution as the parent distribution}
\AttributeTok{  }\FunctionTok{rationale}\KeywordTok{:}\AttributeTok{ the SIA distributions iboth unimodal and right{-}skewed, so a NBD parent is reasonable; more generally, there is evidence of overdispersion relative to a Poisson distribution.}

\KeywordTok{{-}}\AttributeTok{ }\FunctionTok{id}\KeywordTok{:}\AttributeTok{ special\_value\_mid\_risk}
\AttributeTok{  }\FunctionTok{objective}\KeywordTok{:}\AttributeTok{ specify how to model inflation at ages surrounding peak teenage smoking uptake}
\AttributeTok{  }\FunctionTok{decision}\KeywordTok{:}\AttributeTok{ treat ages 12, 13, 14, 19, 20, 21 as a parametrically{-}inflated group, with the extra probability mass there sharing the same distributional shape (mean and dispersion) as the main population }
\AttributeTok{  }\FunctionTok{rationale}\KeywordTok{:}\AttributeTok{ this mid{-}risk subgroup represents early and later smokers surrounding the high{-}risk age subgroup; these ages showed a moderate, secondary elevation in repeated training/test spikeplots of the data}

\KeywordTok{{-}}\AttributeTok{ }\FunctionTok{id}\KeywordTok{:}\AttributeTok{ special\_value\_high\_risk}
\AttributeTok{  }\FunctionTok{objective}\KeywordTok{:}\AttributeTok{ specify how to model inflation at peak teenage smoking uptake ages}
\AttributeTok{  }\FunctionTok{decision}\KeywordTok{:}\AttributeTok{ treat ages 15, 16, 17, 18 as a nonparametrically{-}inflated group, with each age given its own freely{-}estimated extra probability }
\AttributeTok{  }\FunctionTok{rationale}\KeywordTok{:}\AttributeTok{ this high{-}risk subgroup corresponds to frenetic teenage smoking uptake, the age range where the largest number of regular smokers start; these ages showed the largest, most persistent spikes across repeated training/test spikeplots of the data.}

\KeywordTok{{-}}\AttributeTok{ }\FunctionTok{id}\KeywordTok{:}\AttributeTok{ alteration\_operator\_decision}
\AttributeTok{  }\FunctionTok{objective}\KeywordTok{:}\AttributeTok{ determine whether to apply alteration operators}
\AttributeTok{  }\FunctionTok{decision}\KeywordTok{:}\AttributeTok{ do not use alteration operators (A.mix and A.mlm set to NULL)}
\AttributeTok{  }\FunctionTok{rationale}\KeywordTok{:}\AttributeTok{ not needed for this model specification}

\KeywordTok{{-}}\AttributeTok{ }\FunctionTok{id}\KeywordTok{:}\AttributeTok{ deflation\_operator\_decision}
\AttributeTok{  }\FunctionTok{objective}\KeywordTok{:}\AttributeTok{ determine whether to apply deflation operators}
\AttributeTok{  }\FunctionTok{decision}\KeywordTok{:}\AttributeTok{ do not use deflation operators (D.mix and D.mlm set to NULL)}
\AttributeTok{  }\FunctionTok{rationale}\KeywordTok{:}\AttributeTok{ not needed for this model specification}

\KeywordTok{{-}}\AttributeTok{ }\FunctionTok{id}\KeywordTok{:}\AttributeTok{ truncation\_decision}
\AttributeTok{  }\FunctionTok{objective}\KeywordTok{:}\AttributeTok{ determine whether to specify a truncated region}
\AttributeTok{  }\FunctionTok{decision}\KeywordTok{:}\AttributeTok{ do not use truncation (no truncated region specified for SIA)}
\AttributeTok{  }\FunctionTok{rationale}\KeywordTok{:}\AttributeTok{ truncation not appropriate for this response variable}

\KeywordTok{{-}}\AttributeTok{ }\FunctionTok{id}\KeywordTok{:}\AttributeTok{ covariate\_selection}
\AttributeTok{  }\FunctionTok{objective}\KeywordTok{:}\AttributeTok{ select covariates for the linear predictor of parent distribution mean}
\AttributeTok{  }\FunctionTok{decision}\KeywordTok{:}\AttributeTok{ include race2, gender, marital, and educ as covariates}
\AttributeTok{  }\FunctionTok{rationale}\KeywordTok{:}\AttributeTok{ most smoking studies adjust for age, marital status, education and race}

\KeywordTok{{-}}\AttributeTok{ }\FunctionTok{id}\KeywordTok{:}\AttributeTok{ fit\_final\_model}
\AttributeTok{  }\FunctionTok{objective}\KeywordTok{:}\AttributeTok{ fit the final GAITD regression model }
\AttributeTok{  }\FunctionTok{decision}\KeywordTok{:}\AttributeTok{ fit a single vglm that puts together the pieces above {-}{-} response, parent distribution, special{-}value structure (inflation, alteration, deflation, and truncation), and covariates. The covariates only shift the mean of the negative binomial; the dispersion and the mid{-}/high{-}risk inflation probabilities are held fixed across the sample}
\AttributeTok{  }\FunctionTok{rationale}\KeywordTok{:}\AttributeTok{ only the mean is modeled with covariates; the dispersion and inflation probabilities are kept intercept{-}only, since extending covariate effects to them would substantially increase the number of parameters to estimate and make the results harder to interpret}
\end{Highlighting}
\end{Shaded}

}

\end{codelisting}%

Assembling the schema does not require additional analysis from the
authors, only the collection of decisions and rationale already present
in the manuscript and supplementary material.

As a check that this schema is accurate, generating and validating code
from it produces the R script in Listing~\ref{lst-script-sia}, which
reproduces the coefficients reported in Table 2 of Yee et al. (2025).
Since the code was generated without access to the manuscript or the
authors' script, the match shows that the schema records everything
needed to reproduce the analysis. A schema of this kind could therefore
accompany a manuscript as reproducibility material in its own right,
alongside code and data, and a reader can rely on it as an accurate
account of the analysis it documents.

\begin{codelisting}

\caption{\label{lst-script-sia}The R script generated and validated from
the schema in Listing~\ref{lst-heaping}, annotated with the \texttt{id}
of the schema step each block implements.}

\centering{

\begin{Shaded}
\begin{Highlighting}[]
\FunctionTok{library}\NormalTok{(VGAMdata)}
\FunctionTok{library}\NormalTok{(VGAM)}
\FunctionTok{library}\NormalTok{(dplyr)}

\FunctionTok{data}\NormalTok{(}\StringTok{"smqP"}\NormalTok{, }\AttributeTok{package =} \StringTok{"VGAMdata"}\NormalTok{)}

\NormalTok{cols }\OtherTok{\textless{}{-}} \FunctionTok{c}\NormalTok{(}\StringTok{\textquotesingle{}TC30\textquotesingle{}}\NormalTok{, }\StringTok{\textquotesingle{}TCquit\textquotesingle{}}\NormalTok{, }\StringTok{\textquotesingle{}SCA\textquotesingle{}}\NormalTok{, }\StringTok{\textquotesingle{}educ\textquotesingle{}}\NormalTok{, }\StringTok{\textquotesingle{}marital\textquotesingle{}}\NormalTok{,}
          \StringTok{\textquotesingle{}SIA\textquotesingle{}}\NormalTok{, }\StringTok{\textquotesingle{}SEQN\textquotesingle{}}\NormalTok{, }\StringTok{\textquotesingle{}age\textquotesingle{}}\NormalTok{, }\StringTok{\textquotesingle{}gender\textquotesingle{}}\NormalTok{, }\StringTok{\textquotesingle{}race2\textquotesingle{}}\NormalTok{,}
          \StringTok{"passiveSmoke\_job"}\NormalTok{, }\StringTok{"passiveSmoke\_rest"}\NormalTok{,}
          \StringTok{"passiveSmoke\_bar"}\NormalTok{, }\StringTok{"passiveSmoke\_car"}\NormalTok{,}
          \StringTok{"passiveSmoke\_home"}\NormalTok{, }\StringTok{"passiveSmoke\_other"}\NormalTok{,}
          \StringTok{"passiveSmoke\_ecigarettes"}\NormalTok{)}
\NormalTok{df }\OtherTok{\textless{}{-}}\NormalTok{ smqP }\SpecialCharTok{|\textgreater{}} \FunctionTok{select}\NormalTok{(}\FunctionTok{all\_of}\NormalTok{(cols))}

\CommentTok{\# Step 1: complete\_case\_filtering}
\CommentTok{\# Apply complete{-}case filtering to all columns except TC30, TCquit, and SCA}
\NormalTok{cols\_require\_complete }\OtherTok{\textless{}{-}} \FunctionTok{setdiff}\NormalTok{(cols, }\FunctionTok{c}\NormalTok{(}\StringTok{"TC30"}\NormalTok{, }\StringTok{"TCquit"}\NormalTok{, }\StringTok{"SCA"}\NormalTok{))}
\NormalTok{df\_complete }\OtherTok{\textless{}{-}}\NormalTok{ df }\SpecialCharTok{|\textgreater{}}
  \FunctionTok{filter}\NormalTok{(}\FunctionTok{complete.cases}\NormalTok{(}\FunctionTok{pick}\NormalTok{(}\FunctionTok{all\_of}\NormalTok{(cols\_require\_complete))))}

\CommentTok{\# Step 2: age\_restriction}
\CommentTok{\# Restrict analysis to respondents aged less than 80 years}
\NormalTok{df\_age\_restricted }\OtherTok{\textless{}{-}}\NormalTok{ df\_complete }\SpecialCharTok{|\textgreater{}} \FunctionTok{filter}\NormalTok{(age }\SpecialCharTok{\textless{}} \DecValTok{80}\NormalTok{)}

\CommentTok{\# Step 3: data\_quality\_correction}
\CommentTok{\# Set both SCA and SIA to NA where SCA {-} SIA \textless{} 0}
\NormalTok{df\_age\_restricted }\OtherTok{\textless{}{-}}\NormalTok{ df\_age\_restricted }\SpecialCharTok{|\textgreater{}}
  \FunctionTok{mutate}\NormalTok{(}
    \AttributeTok{SCA\_corrected =} \FunctionTok{if\_else}\NormalTok{(}\SpecialCharTok{!}\FunctionTok{is.na}\NormalTok{(SCA) }\SpecialCharTok{\&} \SpecialCharTok{!}\FunctionTok{is.na}\NormalTok{(SIA) }\SpecialCharTok{\&}\NormalTok{ (SCA }\SpecialCharTok{{-}}\NormalTok{ SIA }\SpecialCharTok{\textless{}} \DecValTok{0}\NormalTok{), }\ConstantTok{NA\_integer\_}\NormalTok{, SCA),}
    \AttributeTok{SIA\_corrected =} \FunctionTok{if\_else}\NormalTok{(}\SpecialCharTok{!}\FunctionTok{is.na}\NormalTok{(SCA) }\SpecialCharTok{\&} \SpecialCharTok{!}\FunctionTok{is.na}\NormalTok{(SIA) }\SpecialCharTok{\&}\NormalTok{ (SCA }\SpecialCharTok{{-}}\NormalTok{ SIA }\SpecialCharTok{\textless{}} \DecValTok{0}\NormalTok{), }\ConstantTok{NA\_integer\_}\NormalTok{, SIA)}
\NormalTok{  )}

\CommentTok{\# Step 4: impute\_marital\_young}
\CommentTok{\# Recode marital status to "Never married" for everyone age \textless{} 20}
\NormalTok{df\_age\_restricted }\OtherTok{\textless{}{-}}\NormalTok{ df\_age\_restricted }\SpecialCharTok{|\textgreater{}}
  \FunctionTok{mutate}\NormalTok{(}
    \AttributeTok{marital\_imputed =} \FunctionTok{if\_else}\NormalTok{(age }\SpecialCharTok{\textless{}} \DecValTok{20}\NormalTok{, }\StringTok{"Never married"}\NormalTok{, }\FunctionTok{as.character}\NormalTok{(marital))}
\NormalTok{  )}

\CommentTok{\# Step 5: impute\_education\_young}
\CommentTok{\# Recode education based on age for young respondents}
\NormalTok{df\_age\_restricted }\OtherTok{\textless{}{-}}\NormalTok{ df\_age\_restricted }\SpecialCharTok{|\textgreater{}}
  \FunctionTok{mutate}\NormalTok{(}
    \AttributeTok{educ\_imputed =} \FunctionTok{case\_when}\NormalTok{(}
\NormalTok{      age }\SpecialCharTok{\textless{}} \DecValTok{14} \SpecialCharTok{\textasciitilde{}} \StringTok{"\textless{} 9th grade"}\NormalTok{,}
\NormalTok{      age }\SpecialCharTok{\textgreater{}=} \DecValTok{14} \SpecialCharTok{\&}\NormalTok{ age }\SpecialCharTok{\textless{}=} \DecValTok{16} \SpecialCharTok{\textasciitilde{}} \StringTok{"9{-}11th grade"}\NormalTok{,}
\NormalTok{      age }\SpecialCharTok{\textgreater{}=} \DecValTok{17} \SpecialCharTok{\&}\NormalTok{ age }\SpecialCharTok{\textless{}=} \DecValTok{19} \SpecialCharTok{\textasciitilde{}} \StringTok{"High school"}\NormalTok{,}
      \ConstantTok{TRUE} \SpecialCharTok{\textasciitilde{}} \FunctionTok{as.character}\NormalTok{(educ)}
\NormalTok{    )}
\NormalTok{  )}

\CommentTok{\# Step 6: define\_ever\_smokers\_sample}
\CommentTok{\# Include if: SIA not missing AND (both SCA and TCquit not missing OR SCA missing and TC30 not missing)}
\NormalTok{df\_ever\_smokers }\OtherTok{\textless{}{-}}\NormalTok{ df\_age\_restricted }\SpecialCharTok{|\textgreater{}}
  \FunctionTok{filter}\NormalTok{(}
    \SpecialCharTok{!}\FunctionTok{is.na}\NormalTok{(SIA\_corrected) }\SpecialCharTok{\&}
\NormalTok{    ((}\SpecialCharTok{!}\FunctionTok{is.na}\NormalTok{(SCA\_corrected) }\SpecialCharTok{\&} \SpecialCharTok{!}\FunctionTok{is.na}\NormalTok{(TCquit)) }\SpecialCharTok{|}\NormalTok{ (}\FunctionTok{is.na}\NormalTok{(SCA\_corrected) }\SpecialCharTok{\&} \SpecialCharTok{!}\FunctionTok{is.na}\NormalTok{(TC30)))}
\NormalTok{  )}

\NormalTok{df\_ever\_smokers }\OtherTok{\textless{}{-}} \FunctionTok{droplevels}\NormalTok{(df\_ever\_smokers)}

\CommentTok{\# Step 7: parent\_distribution\_choice}
\CommentTok{\# Adopt negative binomial distribution as the parent distribution}
\NormalTok{parent\_distribution }\OtherTok{\textless{}{-}}\NormalTok{ negbinomial}

\CommentTok{\# Step 8: special\_value\_mid\_risk}
\CommentTok{\# Ages 12, 13, 14, 19, 20, 21 as parametrically{-}inflated group (i.mix)}
\NormalTok{i\_mix\_spec }\OtherTok{\textless{}{-}} \FunctionTok{c}\NormalTok{(}\DecValTok{12}\NormalTok{, }\DecValTok{13}\NormalTok{, }\DecValTok{14}\NormalTok{, }\DecValTok{19}\NormalTok{, }\DecValTok{20}\NormalTok{, }\DecValTok{21}\NormalTok{)}

\CommentTok{\# Step 9: special\_value\_high\_risk}
\CommentTok{\# Ages 15, 16, 17, 18 as nonparametrically{-}inflated group (i.mlm)}
\NormalTok{i\_mlm\_spec }\OtherTok{\textless{}{-}} \FunctionTok{c}\NormalTok{(}\DecValTok{15}\NormalTok{, }\DecValTok{16}\NormalTok{, }\DecValTok{17}\NormalTok{, }\DecValTok{18}\NormalTok{)}

\CommentTok{\# Step 10: alteration\_operator\_decision}
\CommentTok{\# Do not use alteration operators}
\NormalTok{a\_mix\_spec }\OtherTok{\textless{}{-}} \ConstantTok{NULL}
\NormalTok{a\_mlm\_spec }\OtherTok{\textless{}{-}} \ConstantTok{NULL}

\CommentTok{\# Step 11: deflation\_operator\_decision}
\CommentTok{\# Do not use deflation operators}
\NormalTok{d\_mix\_spec }\OtherTok{\textless{}{-}} \ConstantTok{NULL}
\NormalTok{d\_mlm\_spec }\OtherTok{\textless{}{-}} \ConstantTok{NULL}

\CommentTok{\# Step 12: truncation\_decision}
\CommentTok{\# Do not use truncation}
\NormalTok{truncation\_spec }\OtherTok{\textless{}{-}} \ConstantTok{NULL}

\CommentTok{\# Step 13: covariate\_selection}
\CommentTok{\# Include race2, gender, marital, and educ as covariates}
\NormalTok{covariate\_formula }\OtherTok{\textless{}{-}}\NormalTok{ SIA\_corrected }\SpecialCharTok{\textasciitilde{}}\NormalTok{ race2 }\SpecialCharTok{+}\NormalTok{ gender }\SpecialCharTok{+}\NormalTok{ marital\_imputed }\SpecialCharTok{+}\NormalTok{ educ\_imputed}

\CommentTok{\# Step 14: fit\_final\_model}
\CommentTok{\# Fit a single vglm with response, parent distribution, special{-}value structure, and covariates}
\NormalTok{fitted\_model }\OtherTok{\textless{}{-}} \FunctionTok{vglm}\NormalTok{(}
\NormalTok{  covariate\_formula,}
  \AttributeTok{family =} \FunctionTok{gaitdnbinomial}\NormalTok{(}
    \AttributeTok{i.mix =}\NormalTok{ i\_mix\_spec, }\AttributeTok{i.mlm =}\NormalTok{ i\_mlm\_spec,}
    \AttributeTok{a.mix =}\NormalTok{ a\_mix\_spec, }\AttributeTok{a.mlm =}\NormalTok{ a\_mlm\_spec,}
    \AttributeTok{d.mix =}\NormalTok{ d\_mix\_spec, }\AttributeTok{d.mlm =}\NormalTok{ d\_mlm\_spec,}
    \AttributeTok{truncate =}\NormalTok{ truncation\_spec}
\NormalTok{  ),}
  \AttributeTok{data =}\NormalTok{ df\_ever\_smokers}
\NormalTok{)}
\end{Highlighting}
\end{Shaded}

}

\end{codelisting}%

Yee et al. (2025)'s supplementary material reports an investigation
prompted by a reviewer, who suggested splitting the mid-risk group into
two, keeping ages 12--14 as inflated but treating ages 19--21 as altered
instead, since that older subgroup shows a qualitatively different,
declining pattern of uptake. Listing~\ref{lst-ap1921} encodes the
reviewer's suggestion as an alternative file, overriding three steps at
once: \texttt{special\_value\_mid\_risk} narrows the inflated group to
ages 12--14, \texttt{alteration\_operator\_decision} turns on an
alteration group for ages 19--21, and \texttt{covariate\_selection}
drops to an intercept-only model, matching how the authors themselves
computed AIC for this comparison; \texttt{branch:\ single} ties the
three overrides into one refit, rather than crossing them into separate
branches. Generating and validating code from this alternative file
reproduces the AIC of 19,340.4 that Yee et al. (2025) reports. Here the
question was answered because the authors chose to report the refit in
their supplement; with the schema available, any reader who raises such
a question can answer it directly.

\begin{codelisting}

\caption{\label{lst-ap1921}An alternative YAML file specifying a joint
change to the mid-risk inflation, alteration, and covariate steps in the
GAITD schema, treating ages 19--21 as an alteration group rather than
pooling them into the mid-risk inflated group.}

\centering{

\begin{Shaded}
\begin{Highlighting}[]
\FunctionTok{meta}\KeywordTok{:}
\AttributeTok{  }\FunctionTok{type}\KeywordTok{:}\AttributeTok{ alternatives}
\AttributeTok{  }\FunctionTok{branch}\KeywordTok{:}\AttributeTok{ single}
\FunctionTok{nodes}\KeywordTok{:}
\AttributeTok{  }\KeywordTok{{-}}\AttributeTok{ }\FunctionTok{overrides}\KeywordTok{:}\AttributeTok{ special\_value\_mid\_risk}
\AttributeTok{    }\FunctionTok{alternatives}\KeywordTok{:}
\AttributeTok{      }\KeywordTok{{-}}\AttributeTok{ }\FunctionTok{id}\KeywordTok{:}\AttributeTok{ ap1921{-}special{-}value}
\AttributeTok{        }\FunctionTok{decision}\KeywordTok{:}\AttributeTok{ }\StringTok{"treat ages 12, 13, 14 only as a parametrically{-}inflated group, with the extra probability mass there sharing the same distributional shape (mean and dispersion) as the main population; ages 19, 20, 21 are no longer pooled into this group (see the alteration operator decision instead)"}
\AttributeTok{        }\FunctionTok{rationale}\KeywordTok{:}\AttributeTok{ }\StringTok{"ages 19{-}21 mark less frenetic, declining smoking uptake as respondents age out of the high{-}risk window, a qualitatively different pattern from the early uptake at ages 12{-}14, so they are moved out of this pooled inflated group"}
\AttributeTok{  }\KeywordTok{{-}}\AttributeTok{ }\FunctionTok{overrides}\KeywordTok{:}\AttributeTok{ alteration\_operator\_decision}
\AttributeTok{    }\FunctionTok{alternatives}\KeywordTok{:}
\AttributeTok{      }\KeywordTok{{-}}\AttributeTok{ }\FunctionTok{id}\KeywordTok{:}\AttributeTok{ ap1921{-}alteration}
\AttributeTok{        }\FunctionTok{decision}\KeywordTok{:}\AttributeTok{ }\StringTok{"use ages 19, 20, 21 as an alteration group (A.mix) instead of leaving alteration operators unused"}
\AttributeTok{        }\FunctionTok{rationale}\KeywordTok{:}\AttributeTok{ }\StringTok{"ages 19{-}21 mark less frenetic, declining smoking uptake as respondents age out of the high{-}risk window, a qualitatively different pattern from the early uptake at ages 12{-}14, so they are represented as an alteration rather than folded into the same inflated group"}
\AttributeTok{  }\KeywordTok{{-}}\AttributeTok{ }\FunctionTok{overrides}\KeywordTok{:}\AttributeTok{ covariate\_selection}
\AttributeTok{    }\FunctionTok{alternatives}\KeywordTok{:}
\AttributeTok{      }\KeywordTok{{-}}\AttributeTok{ }\FunctionTok{id}\KeywordTok{:}\AttributeTok{ ap1921{-}covariates}
\AttributeTok{        }\FunctionTok{decision}\KeywordTok{:}\AttributeTok{ }\StringTok{"do not include any covariates (intercept{-}only model)"}
\AttributeTok{        }\FunctionTok{rationale}\KeywordTok{:}\AttributeTok{ }\StringTok{"this comparison is about which special{-}value structure fits better, so covariates are left out to compare the two structures on equal footing, the same way the special{-}value structure was first established on an intercept{-}only model before covariates were added"}
\end{Highlighting}
\end{Shaded}

}

\end{codelisting}%

\section{Application: many-analysts football
study}\label{sec-application}

Section~\ref{sec-gaitd} assumed the authors supply a schema alongside
the manuscript, but a schema need not come from the authors. Since it
records decisions in natural language, the schema can be extracted from
the manuscript itself by an LLM for readers to understand and explore
the analysis. In this section, we demonstrate the workflow in the
opposite situation, where no schema is supplied and the workflow serves
as a reproducibility test using the many-analysts football study
(Silberzahn et al., 2018). We choose this study because its
many-analysts design provides replicates of the same analysis:
independent teams analyzed the same data to answer the same question,
each reporting its methodology and estimate.

Silberzahn et al. (2018) invited 29 research groups to conduct the same
data analysis, answering whether referees are more likely to give red
cards to dark-skin-toned players than to light-skin-toned players. The
provided dataset contains two raters' skin color ratings of players'
pictures, the aggregated number of red cards received by a player from a
referee across all matches in that player--referee pair, and other
relevant player, referee, and game characteristics: player birthday,
height, weight, player position, league country, number of victories,
ties, and defeats.

Each team was free to choose its own analysis approach; we take Team 5
as an illustration. They disaggregated the player--referee records into
player--game observations, giving a binary response for whether a player
received a red card, and removed incomplete cases by listwise deletion.
They then fitted a binomial generalized linear mixed model with a logit
link, taking the average of the two raters' skin color ratings as the
covariate of interest, with random intercepts for players and referees
and a random intercept and slope for the referee's country of origin.

Team 5's analysis is only one of the approaches taken: other teams made
different preprocessing decisions, fitted different model families
(linear, logistic, or Poisson regression), specified different fixed and
random effects, or worked in a Bayesian framework. Each team was
required to document its analysis in a report with a publishable
methodology section. All of these are available on the Open Science
Framework.

These reports are not directly usable as input, since the many-analysts
study also covers a second research question and the construction of
confidence intervals. Some teams wrote their final analysis as a set of
edits to the initial one rather than as a self-contained methodology
section. We therefore pre-extract a simplified methodology text for each
team and use it in place of the report.

Each team's text is then passed through the workflow of
Section~\ref{sec-method} to extract the schema and then generate and
validate the corresponding code with Claude Opus 4.5. The Appendix
follows Team 5 through the pipeline to show the input text, the schema,
and the generated script.

To accelerate the evaluation of each script, we manually replace the
standard modeling functions with faster counterparts after the scripts
are generated but before validation: \texttt{lm()} and \texttt{glm()}
with \texttt{fixest::feols()} and \texttt{fixest::feglm()} (Bergé et
al., 2026), and \texttt{lme4::glmer()} (Bates et al., 2015) with
\texttt{glmmTMB::glmmTMB()} (Brooks et al., 2017). These fit the same
models by faster means, using multi-threaded C++ and Template Model
Builder (TMB) respectively, so the substitution changes how the
coefficients are computed, not the coefficients themselves.

Of the 29 teams, the analyses of four teams (Teams 7, 12, 17, and 18)
are too computationally expensive to run inside an iterative
validate-and-repair loop, regardless of whether the schema or generated
code is correct: Team 7 specifies a Dirichlet process profile
regression, Team 12 a zero-inflated Poisson regression, and Teams 17 and
18 MCMC-based hierarchical models. Hence we exclude these four teams
from the evaluation, leaving 25 teams whose analyses we attempt to
reproduce. Across the 25 teams we reproduce, all team scripts were
successfully executed after the validation phase. Notably, two scripts
ran successfully without any validation, whereas the vast majority
reached completion after two or three iterations. The most frequent
correction involved resolving broken and overflowing syntax in comments.

\begin{figure}

\centering{

\pandocbounded{\includegraphics[keepaspectratio]{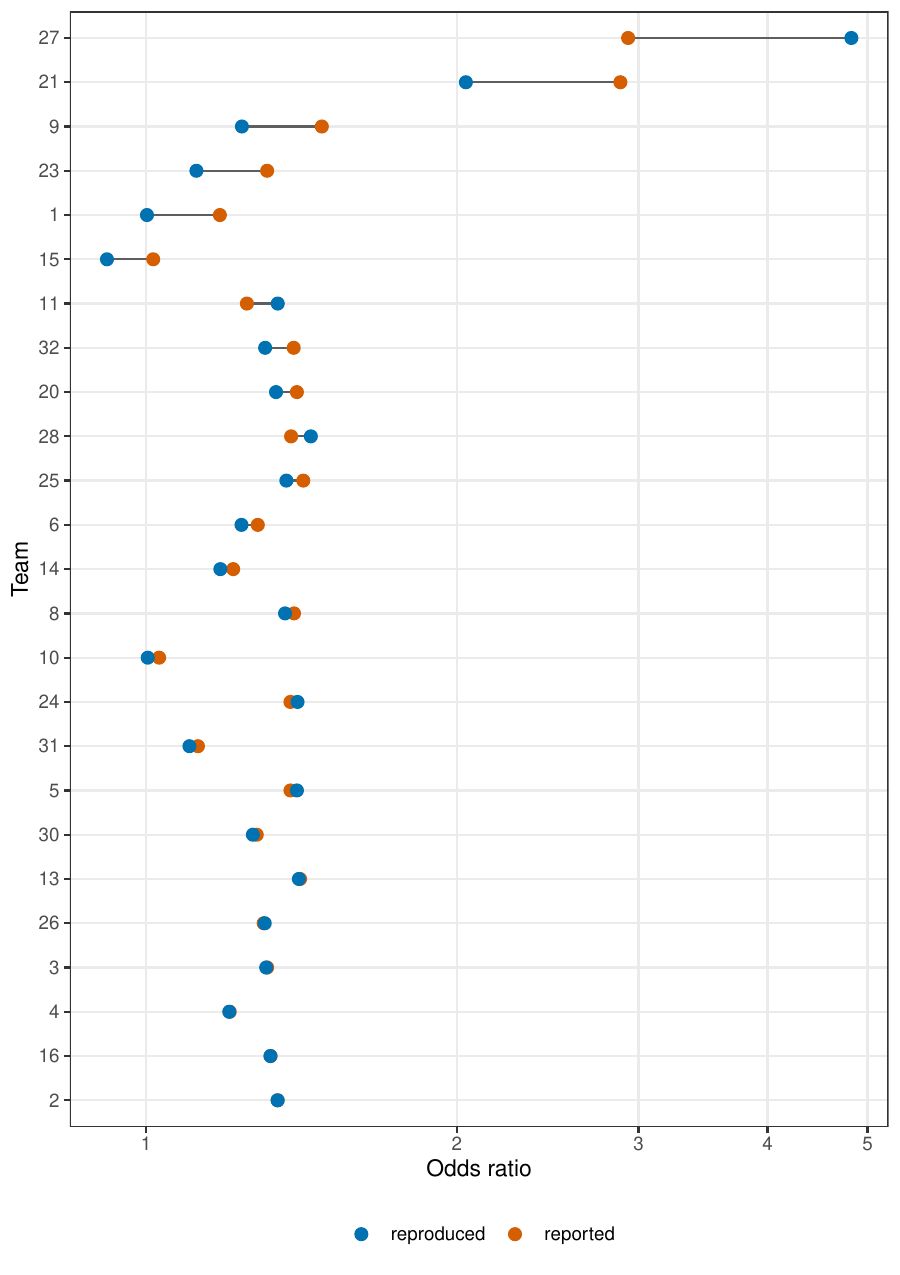}}

}

\caption{\label{fig-res-schema}Comparison of the odds ratio of the
effect of skin color on receiving red cards in football games from our
schema-based reproduction (reproduced) vs.~the effect size reported in
Silberzahn et al. (2018) (reported). Teams are ordered by the absolute
difference between the reproduced and reported odds ratio. The
reproduction recovers the reported odds ratio closely for most teams;
the largest deviations occur for Teams 1, 9, 15, 21, 23, and 27.}

\end{figure}%

Figure~\ref{fig-res-schema} shows the comparison of the estimated effect
sizes, as odds ratios, from our schema workflow (reproduced) and those
reported in Silberzahn et al. (2018) (reported). The odds ratio is well
recovered for most teams, though six show a noticeably larger gap: Teams
1, 9, 15, 21, 23, and 27.

For Teams 1 and 23, the discrepancy originates in our code generation.
When reshaping the data from the player--referee level to the game
level, the schema correctly specifies this step, but the generated code
fails to execute it as intended, and the resulting error passes through
validation silently rather than triggering a repair.

For the remaining four, the reported result could not be regenerated
from the report, though one of them resists diagnosis: for Team 21,
examining both the extracted schema and the generated code turned up
nothing that accounts for the gap. The other three reveal
reproducibility issues in the reports themselves. Team 9's text names
``referee country'' as the covariate while the full report shows
``player country'' is used instead. For Team 27, the authors write that
``a variable for player position was included, as well as a variable for
the league of the player'', which read literally adds only main effects
but might equally be shorthand for interactions, consistent with how the
other covariates entered their initial model. Team 15 is subtler: their
log-linear model includes a skin-tone-by-league interaction, so the
coefficient of interest is defined relative to a parameterization the
report does not state, with SPSS's effect coding averaging it across
leagues and R's default dummy coding fixing it at a reference league.
Setting the \texttt{contrasts} argument in R recovers the reported value
exactly, so the discrepancy is numerical rather than statistical,
although identifying it required manual investigation.

\subsection{Exploring alternatives}\label{sec-explore}

Team 5 arrives at its final model by comparing three random-effects
structures, a set of alternative decisions within a single modeling step
of the kind described in Section~\ref{sec-alternative}. Write
\(y_{ijm}\) for whether player \(i\) received a red card from referee
\(j\) in game \(m\), \(x_i\) for player \(i\)'s skin color rating
averaged over the two raters and rescaled to \([0, 1]\), and \(c(j)\)
for the country of origin of referee \(j\). In each model
\(y_{ijm} \sim \mathrm{Bernoulli}(p_{ijm})\), and the three differ only
in their random effects:

\[
\begin{aligned}
\textbf{GM1:} \quad \mathrm{logit}(p_{ijm}) &= \beta_0 + \beta_1 x_i + u_i + v_j, \\
\textbf{GM2:} \quad \mathrm{logit}(p_{ijm}) &= \beta_0 + \beta_1 x_i + u_i + v_{0j} + v_{1j}\, x_i, \\
\textbf{GM3:} \quad \mathrm{logit}(p_{ijm}) &= \beta_0 + \beta_1 x_i + u_i + v_j + w_{0,c(j)} + w_{1,c(j)}\, x_i,
\end{aligned}
\]

with \(u_i \sim N(0, \sigma^2_{\text{player}})\) throughout,
\(v_j \sim N(0, \sigma^2_{\text{ref}})\) in GM1 and GM3,
\((v_{0j}, v_{1j})^{\top} \sim N(\mathbf{0}, \boldsymbol{\Sigma}_{\text{ref}})\)
in GM2, and
\((w_{0c}, w_{1c})^{\top} \sim N(\mathbf{0}, \boldsymbol{\Sigma}_{\text{country}})\)
in GM3. Model GM1 lets the baseline log-odds vary across players and
referees, GM2 additionally lets the effect of skin color vary across
referees, and GM3, the model Team 5 reports, instead lets both the
baseline and that effect vary across referees' countries of origin.

All three alternatives are specified in a single YAML file, following
the format described in Section~\ref{sec-alternative} and shown in
Listing~\ref{lst-yaml-alternative}. Merging it with Team 5's schema
produces one schema per alternative and yields three validated scripts.
The estimated random and fixed effects in each model match Team 5's
reported values.

\begin{codelisting}

\caption{\label{lst-yaml-alternative}Example of a YAML file specifying
alternatives for the random-effects structure in the many-analysts
analysis of Team 5.}

\centering{

\begin{Shaded}
\begin{Highlighting}[]
\FunctionTok{meta}\KeywordTok{:}
\AttributeTok{  }\FunctionTok{type}\KeywordTok{:}\AttributeTok{ alternatives}
\AttributeTok{  }\FunctionTok{branch}\KeywordTok{:}\AttributeTok{ multi}
\FunctionTok{nodes}\KeywordTok{:}
\KeywordTok{{-}}\AttributeTok{ }\FunctionTok{overrides}\KeywordTok{:}\AttributeTok{ specify\_random\_effects\_structure}
\AttributeTok{  }\FunctionTok{alternatives}\KeywordTok{:}
\AttributeTok{  }\KeywordTok{{-}}\AttributeTok{ }\FunctionTok{id}\KeywordTok{:}\AttributeTok{ gm1}
\AttributeTok{    }\FunctionTok{decision}\KeywordTok{:}\AttributeTok{ a random effect of player and referee}
\AttributeTok{    }\FunctionTok{rationale}\KeywordTok{:}\AttributeTok{ }\StringTok{\textquotesingle{}\textquotesingle{}}
\AttributeTok{  }\KeywordTok{{-}}\AttributeTok{ }\FunctionTok{id}\KeywordTok{:}\AttributeTok{ gm2}
\AttributeTok{    }\FunctionTok{decision}\KeywordTok{:}\AttributeTok{ a random effect of player and referee, and a random effect of skin{-}tone across referees}
\AttributeTok{    }\FunctionTok{rationale}\KeywordTok{:}\AttributeTok{ }\StringTok{\textquotesingle{}\textquotesingle{}}
\AttributeTok{  }\KeywordTok{{-}}\AttributeTok{ }\FunctionTok{id}\KeywordTok{:}\AttributeTok{ gm3}
\AttributeTok{    }\FunctionTok{decision}\KeywordTok{:}\AttributeTok{ a random effect of player, referee, and referees\textquotesingle{} countries of origin, and a random effect of skin tone across referees\textquotesingle{} countries of origin}
\AttributeTok{    }\FunctionTok{rationale}\KeywordTok{:}\AttributeTok{ }\StringTok{\textquotesingle{}\textquotesingle{}}
\end{Highlighting}
\end{Shaded}

}

\end{codelisting}%

\section{Discussion}\label{sec-discussion}

Existing reproducibility practices focus on whether published results
can be reproduced from the code and data, not on the chain of decisions
that produced them. We propose a schema that records those decisions,
along with the rationale for choosing them, in natural language as an
intermediate representation between the manuscript and the code. A
schema can be written by an author or extracted from the manuscript by
an LLM. It can then be turned into an executable script, or expanded
into a multiverse of alternative analyses. For a published GAITD
regression, where the schema was assembled from the manuscript and its
supplement, the resulting script reproduced the reported estimates and
brought a reviewer's alternative specification within reach of any
reader. For the many-analysts football study, where no schema was
supplied, extraction and generation recovered most teams' reported
effects and identified the few where the methodology could have been
described more precisely, inconsistencies that re-running the authors'
own code would not reveal.

A data analysis question admits more than one defensible answer, and the
schema records which one was taken. Arriving at those decisions is a
matter of human judgment, whether the analyst writes the schema directly
or it is extracted from a manuscript in which the decisions were already
made. Once they are determined, generating the code is a comparatively
mechanical translation, which current LLMs perform with reasonable
reliability inside a generation-and-validation loop. Our approach
therefore contrasts with fully automated systems that use multiple LLM
agents to plan, generate, and execute analyses (Hu et al., 2024; Jansen
et al., 2025; Sun et al., 2026), which aim to automate the decisions
themselves. Donoho (2026) describes the risk in such systems as one of
premature autonomy, where the model projects an intent onto the user
rather than responding to what was asked. We do not attempt to automate
the human judgment that goes into making those decisions, but instead
provide a structured way to record them and a workflow to turn them into
code.

Three sources of uncertainty enter this pipeline, and they behave
differently as models improve. The first is implementation, where the
schema states a decision unambiguously and the generated code fails to
carry it out: the scripts for Teams 1 and 23 ran and returned estimates
after reshaping the data incorrectly. The second is extraction, where
the text states a decision clearly to a trained reader and the model
fails to recover it. In Team 5's report, the phrase ``a random effect of
skin-tone across referees' countries of origin'' is read by most
statisticians as implying both a random intercept and a slope, while
read literally it specifies only a slope, and the extraction took the
literal reading rather than the conventional one. Both are epistemic
(reducible) (Hüllermeier \& Waegeman, 2021; Kendall \& Gal, 2017), and
we would expect them to diminish as models continue to improve, rather
than persist as structural limitations of the workflow. The third is
aleatoric (irreducible) and belongs to the source text rather than the
pipeline: where a report is ambiguous or self-contradictory, as in Teams
9, 27, and 15, no model can recover an intent the text does not express.
Team 15 differs in kind: its model is fully specified, and the
discrepancy is numerical, arising because a reparameterization changes
which quantity a coefficient denotes. It remains a reproducibility
issue, since the reported number could be regenerated only by manually
investigating the coding. As extraction and generation improve, what
remains is increasingly uncertainty of this third kind, which the schema
is designed to make visible.

The validate-and-repair loop has two limitations. First, it requires
each candidate script to be executed and diagnosed within an interactive
timeframe, which precludes computationally intensive analyses. For this
reason, the analyses of Teams 7, 12, 17, and 18 were excluded from the
evaluation. Second, for privacy reasons the LLM is given the column
names of the data but not their values, so any data format that the
schema does not specify must be inferred through validation. The birth
dates in the football data, from which player age is derived, illustrate
this: the script for Team 21 initially parsed them as
\texttt{\%Y-\%m-\%d} and adopted the correct \texttt{\%d.\%m.\%Y} only
after a parsing failure, whereas the script for Team 1 took four
attempts to find the correct format. In both cases the format was
recovered by trial and error; had these attempts failed, the pipeline
would either have terminated or propagated the error silently into the
analysis.

We conclude with several directions for future work. Executing scripts
through a background job queue would accommodate analyses too expensive
for an interactive loop without changing the rest of the workflow. The
implementation reported here is an R package, but nothing about the
problem is specific to R, and a general-purpose tool such as a
command-line utility would support analyses written in any language. Our
implementation records the schema as a single YAML file, which gives the
decisions an explicit structure that machines can read. An alternative
is to record each decision in its own Markdown file to trade some of
that structure for flexibility in how a decision is described. Separate
files would further allow decisions to be linked over time, showing how
an analysis developed rather than only where it arrived, and with
attribution to individual researchers such a history could support a
broader research-management system.

\section{Acknowledgments}\label{acknowledgments}

The implementation of the schema is available on CRAN as an R package,
\texttt{tines}. The source code for reproducing the work reported in
this paper is available in the Supplementary Materials and can be found
at: \url{https://github.com/huizezhang-sherry/paper-schema}.

All analyses were carried out in R (R Core Team, 2026), using the
\texttt{tines}, \texttt{tidyverse} (Wickham et al., 2019), \texttt{VGAM}
(Yee, 2025), \texttt{VGAMdata} (Yee \& Gray, 2025), \texttt{lme4} (Bates
et al., 2015), \texttt{glmmTMB} (Brooks et al., 2017; McGillycuddy et
al., 2025), and \texttt{fixest} (Bergé et al., 2026) packages. The
article was created using Quarto (Allaire et al., 2026).

\section*{References}\label{references}
\addcontentsline{toc}{section}{References}

\phantomsection\label{refs}


\begin{CSLReferences}{1}{0}
\bibitem[\citeproctext]{ref-Allaire_Quarto_2026}
Allaire, J. J., Teague, C., Scheidegger, C., Xie, Y., Dervieux, C., \&
Woodhull, G. (2026). \emph{{Quarto}} (Version 1.9) {[}Computer
software{]}. \url{https://doi.org/10.5281/zenodo.5960048}

\bibitem[\citeproctext]{ref-lme4}
Bates, D., Mächler, M., Bolker, B., \& Walker, S. (2015). Fitting linear
mixed-effects models using {lme4}. \emph{Journal of Statistical
Software}, \emph{67}(1), 1--48.
\url{https://doi.org/10.18637/jss.v067.i01}

\bibitem[\citeproctext]{ref-fixest}
Bergé, L. R., Butts, K., \& McDermott, G. (2026). Fast and user-friendly
econometrics estimations: {T}he {R} package fixest. \emph{arXiv Preprint
arXiv:2601.21749}. \url{https://doi.org/10.48550/arXiv.2601.21749}

\bibitem[\citeproctext]{ref-glmmTMB-1}
Brooks, M. E., Kristensen, K., van Benthem, K. J., Magnusson, A., Berg,
C. W., Nielsen, A., Skaug, H. J., Maechler, M., \& Bolker, B. M. (2017).
{glmmTMB} balances speed and flexibility among packages for
zero-inflated generalized linear mixed modeling. \emph{The R Journal},
\emph{9}(2), 378--400. \url{https://doi.org/10.32614/RJ-2017-066}

\bibitem[\citeproctext]{ref-chen2021codex}
Chen, M., Tworek, J., Jun, H., et al. (2021). Evaluating large language
models trained on code. \emph{arXiv Preprint arXiv:2107.03374}.
\url{https://arxiv.org/abs/2107.03374}

\bibitem[\citeproctext]{ref-donoho2026}
Donoho, D. (2026). Discussion of {``}LAMBDA: A large model based data
agent{''}. \emph{Journal of the American Statistical Association},
\emph{121}(553), 14--16.
\url{https://doi.org/10.1080/01621459.2025.2574396}

\bibitem[\citeproctext]{ref-hu_automated_2024}
Hu, S., Lu, C., \& Clune, J. (2024, October). \emph{Automated {Design}
of {Agentic} {Systems}}.
\url{https://openreview.net/forum?id=t9U3LW7JVX}

\bibitem[\citeproctext]{ref-huang_executionbased_2022}
Huang, J., Wang, C., Zhang, J., Yan, C., Cui, H., Inala, J. P., Clement,
C., \& Duan, N. (2022). Execution-based {Evaluation} for {Data}
{Science} {Code} {Generation} {Models}. In E. Dragut, Y. Li, L. Popa, S.
Vucetic, \& S. Srivastava (Eds.), \emph{Proceedings of the {Fourth}
{Workshop} on {Data} {Science} with {Human}-in-the-{Loop} ({Language}
{Advances})} (pp. 28--36). Association for Computational Linguistics.
\url{https://doi.org/10.18653/v1/2022.dash-1.5}

\bibitem[\citeproctext]{ref-hullermeier2021}
Hüllermeier, E., \& Waegeman, W. (2021). Aleatoric and epistemic
uncertainty in machine learning: An introduction to concepts and
methods. \emph{Machine Learning}, \emph{110}(3), 457--506.

\bibitem[\citeproctext]{ref-jansen2025}
Jansen, J. A., Manukyan, A., Khoury, N. A., \& Akalin, A. (2025).
Leveraging large language models for data analysis automation.
\emph{PLOS ONE}, \emph{20}(2), e0317084.
\url{https://doi.org/10.1371/journal.pone.0317084}

\bibitem[\citeproctext]{ref-jiang2026survey}
Jiang, J., Wang, F., Shen, J., Kim, S., \& Kim, S. (2026). A survey on
large language models for code generation. \emph{ACM Transactions on
Software Engineering and Methodology}, \emph{35}(2), 1--72.

\bibitem[\citeproctext]{ref-kendall2017}
Kendall, A., \& Gal, Y. (2017). What uncertainties do we need in
bayesian deep learning for computer vision? \emph{Advances in Neural
Information Processing Systems 30 (NeurIPS 2017)}, 5574--5584.

\bibitem[\citeproctext]{ref-jupyter}
Kluyver, T., Ragan-Kelley, B., Pérez, F., Granger, B., Bussonnier, M.,
Frederic, J., Kelley, K., Hamrick, J., Grout, J., Corlay, S., Ivanov,
P., Avila, D., Abdalla, S., Willing, C., \& team, J. development.
(2016). Jupyter notebooks - a publishing format for reproducible
computational workflows. In F. Loizides \& B. Scmidt (Eds.),
\emph{Positioning and power in academic publishing: Players, agents and
agendas} (pp. 87--90). IOS Press.
\url{https://eprints.soton.ac.uk/403913/}

\bibitem[\citeproctext]{ref-lai_ds1000_2023}
Lai, Y., Li, C., Wang, Y., Zhang, T., Zhong, R., Zettlemoyer, L., Yih,
W.-T., Fried, D., Wang, S., \& Yu, T. (2023). {DS}-1000: {A} {Natural}
and {Reliable} {Benchmark} for {Data} {Science} {Code} {Generation}.
\emph{Proceedings of the 40th {International} {Conference} on {Machine}
{Learning}}, 18319--18345.
\url{https://proceedings.mlr.press/v202/lai23b.html}

\bibitem[\citeproctext]{ref-Leisch2002}
Leisch, F. (2002). Sweave: Dynamic generation of statistical reports
using literate data analysis. In W. Härdle \& B. Rönz (Eds.),
\emph{Compstat 2002 --- proceedings in computational statistics} (pp.
575--580). Physica-Verlag GmbH.

\bibitem[\citeproctext]{ref-glmmTMB-2}
McGillycuddy, M., Warton, D. I., Popovic, G., \& Bolker, B. M. (2025).
Parsimoniously fitting large multivariate random effects in {glmmTMB}.
\emph{Journal of Statistical Software}, \emph{112}(1), 1--19.
\url{https://doi.org/10.18637/jss.v112.i01}

\bibitem[\citeproctext]{ref-peng_reproducible_2011}
Peng, R. D. (2011). Reproducible {Research} in {Computational}
{Science}. \emph{Science}, \emph{334}(6060), 1226--1227.
\url{https://doi.org/10.1126/science.1213847}

\bibitem[\citeproctext]{ref-R}
R Core Team. (2026). \emph{R: A language and environment for statistical
computing}. R Foundation for Statistical Computing.
\url{https://doi.org/10.32614/R.manuals}

\bibitem[\citeproctext]{ref-sarma2023}
Sarma, A., Kale, A., Moon, M. J., Taback, N., Chevalier, F., Hullman,
J., \& Kay, M. (2023). Multiverse: Multiplexing alternative data
analyses in r notebooks. \emph{Proceedings of the 2023 CHI Conference on
Human Factors in Computing Systems}.
\url{https://doi.org/10.1145/3544548.3580726}

\bibitem[\citeproctext]{ref-shah_automating_2026}
Shah, S. M. H., Hopfgartner, F., \& Bleier, A. (2026). \emph{Automating
{Computational} {Reproducibility} in {Social} {Science}: {Comparing}
{Prompt}-{Based} and {Agent}-{Based} {Approaches}}.
\url{https://doi.org/10.1145/3774905.3795485}

\bibitem[\citeproctext]{ref-silberzahn2018}
Silberzahn, R., Uhlmann, E. L., Martin, D. P., Anselmi, P., Aust, F.,
Awtrey, E., Bahník, Š., Bai, F., Bannard, C., Bonnier, E., Carlsson, R.,
Cheung, F., Christensen, G., Clay, R., Craig, M. A., Dalla Rosa, A.,
Dam, L., Evans, M. H., Flores Cervantes, I., \ldots{} Nosek, B. A.
(2018). Many Analysts, One Data Set: Making Transparent How Variations
in Analytic Choices Affect Results. \emph{Advances in Methods and
Practices in Psychological Science}, \emph{1}(3), 337--356.
\url{https://doi.org/10.1177/2515245917747646}

\bibitem[\citeproctext]{ref-simonsohn2020}
Simonsohn, U., Simmons, J. P., \& Nelson, L. D. (2020). Specification
curve analysis. \emph{Nature Human Behaviour}, \emph{4}(11), 1208--1214.
\url{https://doi.org/10.1038/s41562-020-0912-z}

\bibitem[\citeproctext]{ref-steegen2016}
Steegen, S., Tuerlinckx, F., Gelman, A., \& Vanpaemel, W. (2016).
Increasing transparency through a multiverse analysis.
\emph{Perspectives on Psychological Science}, \emph{11}(5), 702--712.
\url{https://doi.org/10.1177/1745691616658637}

\bibitem[\citeproctext]{ref-sun_lambda_2026}
Sun, M., Han, R., Jiang, B., Qi, H., Sun, D., Yuan, Y., \& Huang, J.
(2026). {LAMBDA}: {A} {Large} {Model} {Based} {Data} {Agent}.
\emph{Journal of the American Statistical Association}, \emph{121},
1--13. \url{https://doi.org/10.1080/01621459.2025.2510000}

\bibitem[\citeproctext]{ref-hdi}
United Nations Development Programme. (2025). \emph{Human development
report 2025}. United Nations Development Programme.
\url{https://hdr.undp.org/}

\bibitem[\citeproctext]{ref-wickham2019tidyverse}
Wickham, H., Averick, M., Bryan, J., Chang, W., McGowan, L. D.,
François, R., Grolemund, G., Hayes, A., Henry, L., Hester, J., Kuhn, M.,
Pedersen, T. L., Miller, E., Bache, S. M., Müller, K., Ooms, J.,
Robinson, D., Seidel, D. P., Spinu, V., \ldots{} Yutani, H. (2019).
Welcome to the tidyverse. \emph{Journal of Open Source Software},
\emph{4}(43), 1686. \url{https://doi.org/10.21105/joss.01686}

\bibitem[\citeproctext]{ref-wrobel_partnering_2024}
Wrobel, J., Hector, E. C., Crawford, L., McGowan, L. D., Silva, N. da,
Goldsmith, J., Hicks, S., Kane, M., Lee, Y., Mayrink, V., Paciorek, C.
J., Usher, T., \& Wolfson, J. (2024). Partnering {With} {Authors} to
{Enhance} {Reproducibility} at {JASA}. \emph{Journal of the American
Statistical Association}, \emph{119}(546), 795--797.
\url{https://doi.org/10.1080/01621459.2024.2340557}

\bibitem[\citeproctext]{ref-xie2018rmarkdown}
Xie, Y., Allaire, J. J., \& Grolemund, G. (2018). \emph{R markdown: The
definitive guide}. Chapman; Hall/CRC.
\url{https://bookdown.org/yihui/rmarkdown}

\bibitem[\citeproctext]{ref-VGAM}
Yee, T. W. (2025). \emph{{VGAM}: Vector generalized linear and additive
models}. \url{https://CRAN.R-project.org/package=VGAM}

\bibitem[\citeproctext]{ref-yee2025heaping}
Yee, T. W., Frigau, L., \& Ma, C. (2025). Heaping and seeping, GAITD
regression and doubly constrained reduced-rank vector generalized linear
models in smoking studies. \emph{The Annals of Applied Statistics},
\emph{19}(4), 3045--3070.

\bibitem[\citeproctext]{ref-VGAMdata}
Yee, T. W., \& Gray, J. (2025). \emph{{VGAMdata}: Data supporting the
{VGAM} package}. \url{https://CRAN.R-project.org/package=VGAMdata}

\end{CSLReferences}
\end{document}